# Co-Occurring of Object Detection and Identification towards unlabeled object discovery

BINAY KUMAR SINGH

Department of Computer Science, University of Central Florida, United States of America

NIELS DA VITORIA LOBO

Department of Computer Science, University of Central Florida, United States of America

In this paper, we propose a novel deep learning based approach for identifying co-occurring objects in conjunction with base objects in multilabel object categories. Nowadays, with the advancement in computer vision based techniques we need to know about co-occurring objects with respect to base object for various purposes. The pipeline of the proposed work is composed of two stages: in the first stage of the proposed model we detect all the bounding boxes present in the image and their corresponding labels, then in the second stage we perform co-occurrence matrix analysis. In co-occurrence matrix analysis, we set base classes based on the maximum occurrences of the labels and build association rules and generate frequent patterns. These frequent patterns will show base classes and their corresponding co-occurring classes. We performed our experiments on two publicly available datasets: Pascal VOC and MS-COCO. The experimental results on public benchmark dataset is reported in Sec 4. Further we extend this work by considering all frequently objects as unlabeled and what if they are occluded as well.

**Additional Keywords and Phrases:** Deep Learning framework, co-occurring object analysis,

## 1 INTRODUCTION

Several research has been done in deep learning based models for multilabel classification[3] based on the object detectors such as RCNN based detectors where backbone network is based on Convolutional Neural Network(CNN) and transformer based detectors. With the introduction of transformers based model in vision related tasks performance of DNN based model improved up to the label of a human being. These networks can perform a variety of tasks in computer vision such as object detection, segmentation, object classification and so on. This research work is motivated by the growing demand for robots used in navigation and scene understanding. Multilabel classification is a typical problem in computer vision area which is different than classical multiclass classification where mostly all images are classified into one class only. Here in this paper we are extending the problem of multilabel multiclass classification where we also want to find all the co-occurring object classes frequently occurring with base class. So it is important here to differentiate between base class and co-occurring class. In this problem setting, when we say base class that means we are referring to objects that appeared maximum number of times in the whole dataset, and co-occurring objects means these objects frequently occurred with base class more than a threshold value. Object when frequently occur with other objects chances are high that one object is related to another object or when objects share some common context, they tend to appear more closely to each other.

Our visual cortex, especially para hippocampal cortex is responsible for representing strong contextual associations ref [1] with other objects or locations in co-occurring objects. By using contextual knowledge generated from our surroundings we people try to identify objects, for example, looking at a computer desk our visual cortex captures all items on the desk like monitor, keyboard, mouse, and other stationary items. One interesting implementation of the proposed model is in security and surveillance. Consider a scenario at a crowded airport, one suspect was found now the security team task is to find any other suspect related with this suspect, by using this approach it will be easy because the model will identify other co-occurring person related with this person. Some research towards learning co-occurrence statistics of object representation is done with the help of vision and language models ref [2], but in our research we are not referring to any language models algorithms such as object2vec or word2vec.

In this research work, we seek to develop a deep learning based model that can find all base class objects and their corresponding co-occurring objects and report the co-occurrence statistics in respect to base class objects. The proposed method can be described in a two stage pipeline as follows: the first stage of the pipeline deals with predicting class labels. To do so we used deep convolutional neural networks based feature extractor and classifier for multiclass multilabel object detection and co-occurrence matrix that generates frequently occurring objects corresponding to each base classes.

Our selection of object detection models are limited in the fact that here we need a model which can detection multiple objects in the images. Models such as any classification based models are ruled out because these models performed well on single object detection/classification, while models based on detection and segmentation performed well.

In summary, our contribution in this paper is four folded listed below.

- We proposed a new method for co-occurring object detection and identification.
- We used feature extractors followed by classifier for multilabel multiclass classification and co-occurrence matrix analysis for frequently occurring objects in terms of base classes and co-occurring classes.
- We formulate a novel solution by using contextuality and compositionality when co-occurring object are unknown and when they are partially occluded.
- Finally evaluate the proposed method on various evaluation metrics on two publicly available dataset.

## 2 RELATED WORK

The next subsections provide details on background information required for this research work.

### 2.1 Multilabel Object Detectors

Multilabel multiclass object detection is a crucial task in computer vision and machine learning, aiming to identify and classify multiple objects within an image while allowing for the possibility of multiple objects belonging to different classes to be present simultaneously in one single image. This complex task finds applications in various domains such as autonomous driving, surveillance, medical imaging, and more. There are many object detectors available in which the family of object detectors based on Region based Convolutional Neural Network(RCNN) is mostly used by researchers due to its computation efficiency and better performance measurers, metrics such as average precision(AP), mean average precision(mAP) over its counterparts and are of two variations- single stage detector (YOLO versions), SSD, RetinaNet, EfficientDet, etc., and two stage detector- RCNN, fast RCNN, faster RCNN, mask RCNN, Detectron and some detectors based on transformer architecture. For any object detector the key components are feature extraction, object detection, classification and labelling, and post-processing. While developing any object detector there are some challenges researchers faces, for example, label ambiguity, data imbalances, scalability, and real-time performance.

### 2.2 Co-occurrence Object Prediction

One of the challenges in multilabel multiclass problem is huge, annotated data that leads to problems such learning about co-occurrence of objects, which we are dealing in this research paper. In ref [5], authors presents a semi-supervised based learning approach where for a trained network labels are used for inferring unlabeled images. Nowadays, in data mining also, researchers use frequency pattern analysis for identifying, ref [6]. In frequency pattern analysis a tree based approach is used formerly known as frequency pattern growth, ref [7] for mining data elements.

ALGORITHM 1: Co-occurrence object Prediction

Co-occurObjPredict(A);

**Input** : A: Contains multilabel dataset

**Output**: Table showing base object and corresponding co-occurring object

**Procedure**

1. Perform feature extraction on each image of the dataset
2. Generate bounding boxes and class labels for each object
3. Define base classes and co-occurring classes
4. Perform co-occurring matrix analysis and find base classes and corresponding co-occurring objects
5. Report table containing base classes and their corresponding co-occurring object classes

## 3 THE PROPOSED METHOD

### 3.1 Overview the Proposed Model

Our proposed approach is presented in Figure 1 consists of multi-label multi class object classification and co-occurrence object prediction. The proposed approach comprised of two stages. Given a target DCNN architecture for object detection our goal is to generate a set of class labels for each object present in the image.

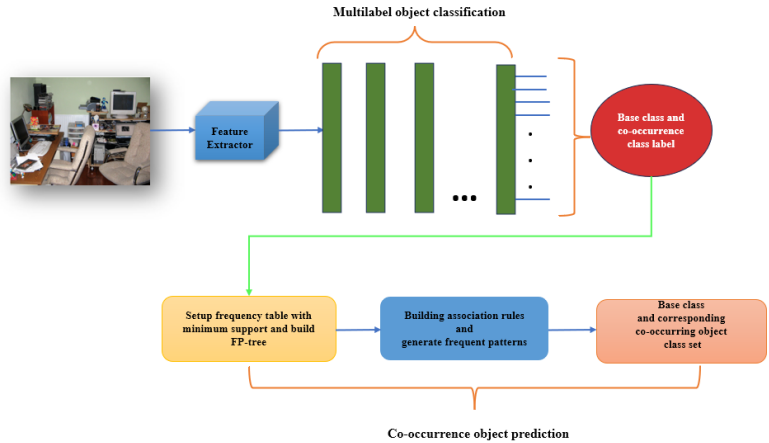

Figure 1: A unified feed-forward framework for co-occurrence object prediction. For each image in a dataset, the DCNN backbone feature extractor with multilabel object classifier generates a set of bounding boxes and labels that is further processed for base class and co-occurrence class generation.

The first stage consists of multilabel multiclass object labelling and classification and in the second stage we perform co-occurrence object analysis. From first stage we detect all the labels of an image by generating a set of bounding boxes and labeling them for multilabel classification, the learnable parameters are only associated with this part of the network. For this purpose, we used two state-of-the-art object detectors based on Faster RCNN and transformers. In faster RCNN based architecture we used two different backbone networks- ResNet50 and mobilenetv3. We fine tune the pre-trained architectures based on the datasets.

In the second stage, from the multilabel multiclass data we identify base classes and co-occurring classes. Base classes as defined earlier which must be present maximum number of times in the dataset, and co-occurring classes are those classes that appear frequently (greater than a minimum support value). After finding base classes by finding the frequency of all the labels in all the image set, we build association rules. These association rules will help in generate frequent patterns. The frequent patterns will contain base classes and the corresponding co-occurring classes.

### 3.2  Co-Occurrence Label Prediction with Base Classes

The co-occurrence object label prediction [4] takes both base class and frequently occurring at once and set up frequency table, here threshold $t = 0.5$ is set that filters out infrequently occurring objects. In the next step we build frequency pattern tree that will build association rules and finally generate frequent patterns. These frequent patterns contains base class and their corresponding co-occurring class and the number of occurrences of these co-occurring classes in respect to base class.

## 4 EXPERIMENTS AND RESULTS

As per the proposed methodologies, our experiments consist of two stages: in the first stage we identify all the labels, to do so, we used a set of state of the art object detectors based on RCNN(Region based Convolutional Neural Network) and transformer based deep neural network. In co-occurring object detection and identification, we must have multi label multi classification dataset, which was challenging, because most of the datasets are multiclass classification so we decided to use two publicly available dataset for multilabel multiclass problem, Pascal VOC and MS-COCO, although not all images contain more than one label in both the datasets, details of each of the dataset is presented below. We take Faster RCNN based on two backbones: ResNet50 and mobilenetv2 pretrained on ImageNet. We replace the last fully-connected(classification layer) with 20 output units for Pascal VOC and 80 output units for MS-COCO. We retrain the model and extract all the object labels for each image. We use optimizer as 'Adam' and loss function as binary cross entropy with logits. We run the experiment for 30 epochs and learning rate is 0.01.

For Pascal VOC we used 'train' set for training and 'val' set for evaluation. The metric used here is average precision and mAP(mean Average Precision). In this dataset, we performed data augmentations to obtain better precision of the model. Transformations such as Random Choice, Random Horizontal Flip, Random Rotation and Normalization is used and resized the original image of different resolution sizes to (300x300) and fed into the first component of the proposed method. The first component is based on Faster RCNN and transformer based models. In FasterRCNN based object detector, the backbone network is based on ResNet50 and mobilenetv3 where we modified the last cls_score as per the datasets and predicted bounding boxes and class labels for the set of objects present in the images and accumulated the whole set of image labels for further processing. In the next stage of the pipeline we set threshold value for base object class and co-occurring object classes. After that, by going through a series of steps(building association rules and generating frequent patterns) we generated base classes and their corresponding co-occurring classes.

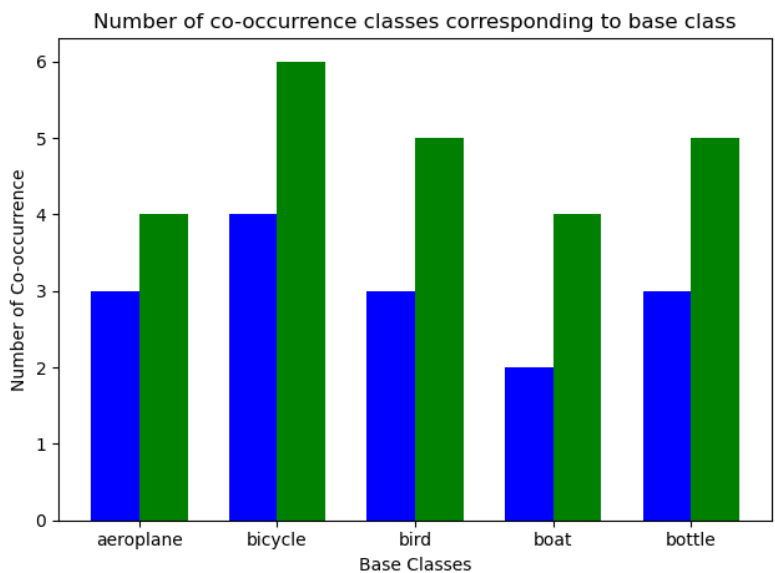

Figure 2: This chart represents base classes and their corresponding number of co-occurrence classes. Here, blue tick refers to model1(Faster RCNN backbone with ResNet50) and green tick refers to model 2 (Faster RCNN backbone with mobilenetv2).

As shown in Fig. 2 we can infer that model 1 when used backbone as ResNet50 performed better over model 2 based on mobilenetv3.

## 4.1 Results

Here, Table 1 shows performance of two baseline models- backbone on ResNet50 and Mobilenetv3. When ResNet50 backbone is used we see that AP and mAP is here than mobilenetv3.

Table 1: Model performance on various parameters.

| Dataset | Models | AP | mAP |
| --- | --- | --- | --- |
| Pascal VOC | ResNet50 | 71.5 | 68.3 |
| | Mobilenetv3 | 65.2 | 60.1 |

## 5 CONCLUSION

We have presented a new framework for frequently occurring objects corresponding to base classes. We proposed the method in two stages where in the first stage we locate all the objects present in the image and their corresponding labels, and in the second stage we find all base classes and their corresponding co-occurring classes set by a threshold value. Our experiments on two different datasets on two different deep network architecture is presented. In the future, we plan to extend this work and will consider co-occurring classes as unknown and occluded.